\documentclass[letterpaper]{article} 

\usepackage{aaai24}  

\usepackage{times}  
\usepackage{helvet}  
\usepackage{courier}  
\usepackage[hyphens]{url}  
\usepackage{graphicx} 
\usepackage{natbib}  
\usepackage{caption} 
\usepackage{algorithm}
\usepackage{algorithmic}

\newcommand{\openeds}{OpenEDS2020}

\newcommand{\aea}{AEA}
\newcommand{\mpiigaze}{MPIIGaze}
\newcommand{\mpiifacegaze}{MPIIFaceGaze}

\usepackage[dvipsnames]{xcolor}

\definecolor{Black}{rgb}{0.0, 0.0, 0.0}

\definecolor{DarkGreen}{rgb}{0.10, 0.55, 0.10}
\definecolor{DeepSkyBlue3}{rgb}{0.0, 0.686, 0.843}
\definecolor{DarkTurquoise}{rgb}{0.0, 0.843, 0.843}
\definecolor{Cyan3}{rgb}{0.0, 0.843, 0.686}
\definecolor{LightSeaGreen}{rgb}{0.0, 0.686, 0.686} 

\definecolor{RoyalBlue}{rgb}{0.20, 0.60, 0.86}
\definecolor{DeepSkyBlue}{rgb}{0.0, 0.686, 1.}

\definecolor{DodgerBlue}{rgb}{0.0, 0.529, 1.} 
\definecolor{DodgerBlue2}{rgb}{0.0, 0.3725, 1.}
\definecolor{DodgerBlue3}{rgb}{0.0, 0.3725, 0.843}
\definecolor{DarkCyan}{rgb}{0.0, 0.54, 0.54}

\definecolor{Gray}{gray}{0.9}
\definecolor{ChromeYellow}{rgb}{1.0, 0.65, 0.0}

\definecolor{Gold}{rgb}{1.0, 0.843, 0.0}

\definecolor{Crimson}{rgb}{0.86, 0.08, 0.24}
\definecolor{IndianRed}{rgb}{1.0, 0.373, 0.529}

\definecolor{SunsetOrange}{rgb}{0.98, 0.37, 0.33}
\definecolor{DarkOrange}{rgb}{1.0, 0.529, 0.}
\ifx \submission \undefined
\newcommand{\hy}[1]{\textcolor{Black}{#1}}

\else
\newcommand{\hy}[1]{{#1}}

\fi

\usepackage{booktabs}
\usepackage{animate}
\usepackage{tabularx}

\newcommand{\etal}{\textit{et al}.}

\newcommand{\eg}{\textit{e}.\textit{g}.} 

\newcommand{\numColumns}{3}
\newcommand{\columnSpacing}{0.25em}
\newcommand{\animationNotes}{
\textit{Best viewed in Acrobat Reader; click images to play animations.}
}

\usepackage{amsmath} 

\usepackage{amssymb}

\usepackage{newfloat}
\usepackage{listings}
\DeclareCaptionStyle{ruled}{labelfont=normalfont,labelsep=colon,strut=off} 
\floatstyle{ruled}
\newfloat{listing}{tb}{lst}{}
\floatname{listing}{Listing}
\title{GazeGen: Gaze-Driven User Interaction for Visual Content Generation}
\author{
    He-Yen Hsieh\textsuperscript{\rm 1}, 
    Ziyun Li\textsuperscript{\rm 2}, 
    Sai Qian Zhang\textsuperscript{\rm 2,3}, 
    Wei-Te Mark Ting\textsuperscript{\rm 1}, 
    Kao-Den Chang\textsuperscript{\rm 1}, \\
    Barbara De Salvo\textsuperscript{\rm 2},
    Chiao Liu\textsuperscript{\rm 2}, 
    H. T. Kung\textsuperscript{\rm 1}
}
\affiliations{
    \textsuperscript{\rm 1}Harvard University \quad
    \textsuperscript{\rm 2}Reality Labs Research, Meta \quad
    \textsuperscript{\rm 3}New York University
}

\usepackage{bibentry}

\begin{document}


\twocolumn[{%
\renewcommand\twocolumn[1][]{#1}%
\maketitle
\vspace{-10mm}
\begin{center}
    \centering
    \includegraphics[width=0.98\linewidth]{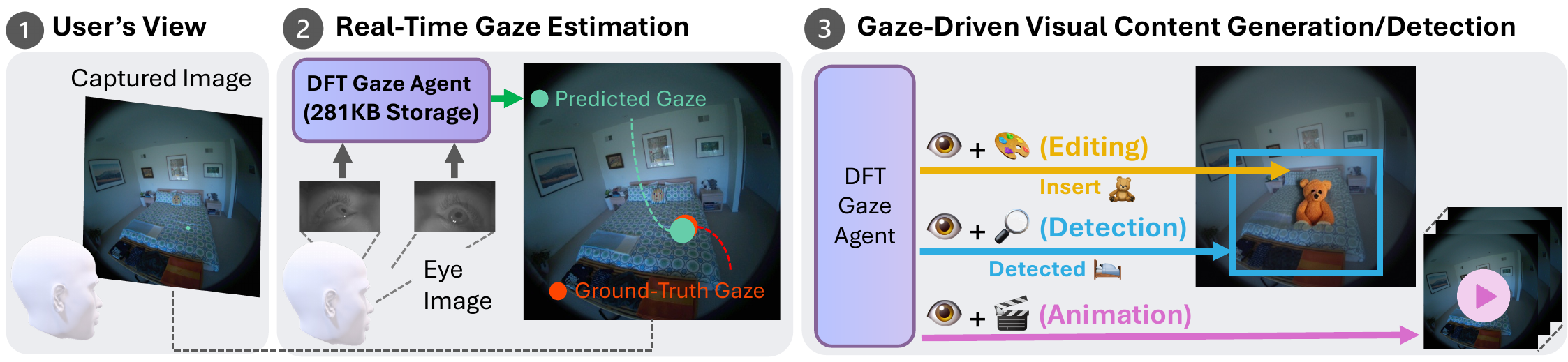}
    \captionof{figure}{
    \hy{GazeGen. (1) \textit{User's View}: Overview of the user's view, setting the context for gaze estimation (input: user's eye images) and visual editing (inputs: user's view and predicted gaze point). (2) \textit{Real-Time Gaze Estimation}: The DFT Gaze Agent (281KB storage) predicts the user's gaze point (\textcolor{ForestGreen}{green}) aligned with the ground-truth gaze (\textcolor{Red}{red}). (3) \textit{Gaze-Driven Visual Content Generation/Detection}: Predicted gaze is used for editing (\includegraphics[height=0.8em]{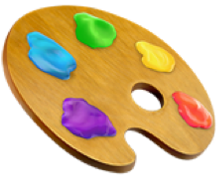}) objects, detecting (\includegraphics[height=0.8em]{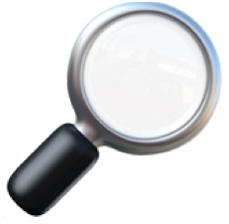}) objects, or creating animations (\includegraphics[height=0.8em]{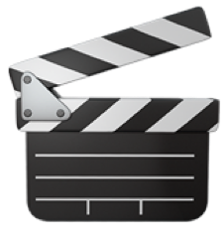}) based on the user's focus (\includegraphics[height=0.8em]{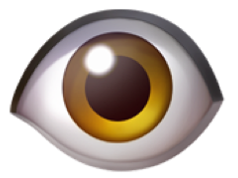}). GazeGen sets a new standard for gaze-driven visual content generation, enhancing user experience and positioning users as visual creators.
    }
    }
    \label{fig:teaser}
\end{center}
}]


\begin{abstract}
\hy{
We present GazeGen, a user interaction system that generates visual content (images and videos) for locations indicated by the user's eye gaze. GazeGen allows intuitive manipulation of visual content by targeting regions of interest with gaze. Using advanced techniques in object detection and generative AI, GazeGen performs gaze-controlled image adding/deleting, repositioning, and surface style changes of image objects, and converts static images into videos.
Central to GazeGen is the DFT Gaze (Distilled and Fine-Tuned Gaze) agent, an ultra-lightweight model with only 281K parameters, performing accurate real-time gaze predictions tailored to individual users' eyes on small edge devices. GazeGen is the first system to combine visual content generation with real-time gaze estimation, made possible exclusively by DFT Gaze. This real-time gaze estimation enables various visual content generation tasks, all controlled by the user's gaze. The input for DFT Gaze is the user's eye images, while the inputs for visual content generation are the user's view and the predicted gaze point from DFT Gaze.
To achieve efficient gaze predictions, we derive the small model from a large model (10x larger) via novel knowledge distillation and personal adaptation techniques. We integrate knowledge distillation with a masked autoencoder, developing a compact yet powerful gaze estimation model. This model is further fine-tuned with Adapters, enabling highly accurate and personalized gaze predictions with minimal user input. DFT Gaze ensures low-latency and precise gaze tracking, supporting a wide range of gaze-driven tasks in AR environments.
Leveraging these precise gaze predictions, GazeGen facilitates visual content generation through diffusion processes, allowing users to intuitively manipulate visual content by targeting regions with their gaze. Additionally, it enables real-time object detection by focusing on specific regions indicated by the user's gaze, improving responsiveness.
We validate the performance of DFT Gaze on AEA and OpenEDS2020 benchmarks, demonstrating low angular gaze error and low latency on the edge device (Raspberry Pi 4). Furthermore, we describe applications of GazeGen, illustrating its versatility and effectiveness in various usage scenarios.
}
\end{abstract}


\vspace{-5mm}
\begin{figure*}
    \centering
    \includegraphics[width=0.9\linewidth]{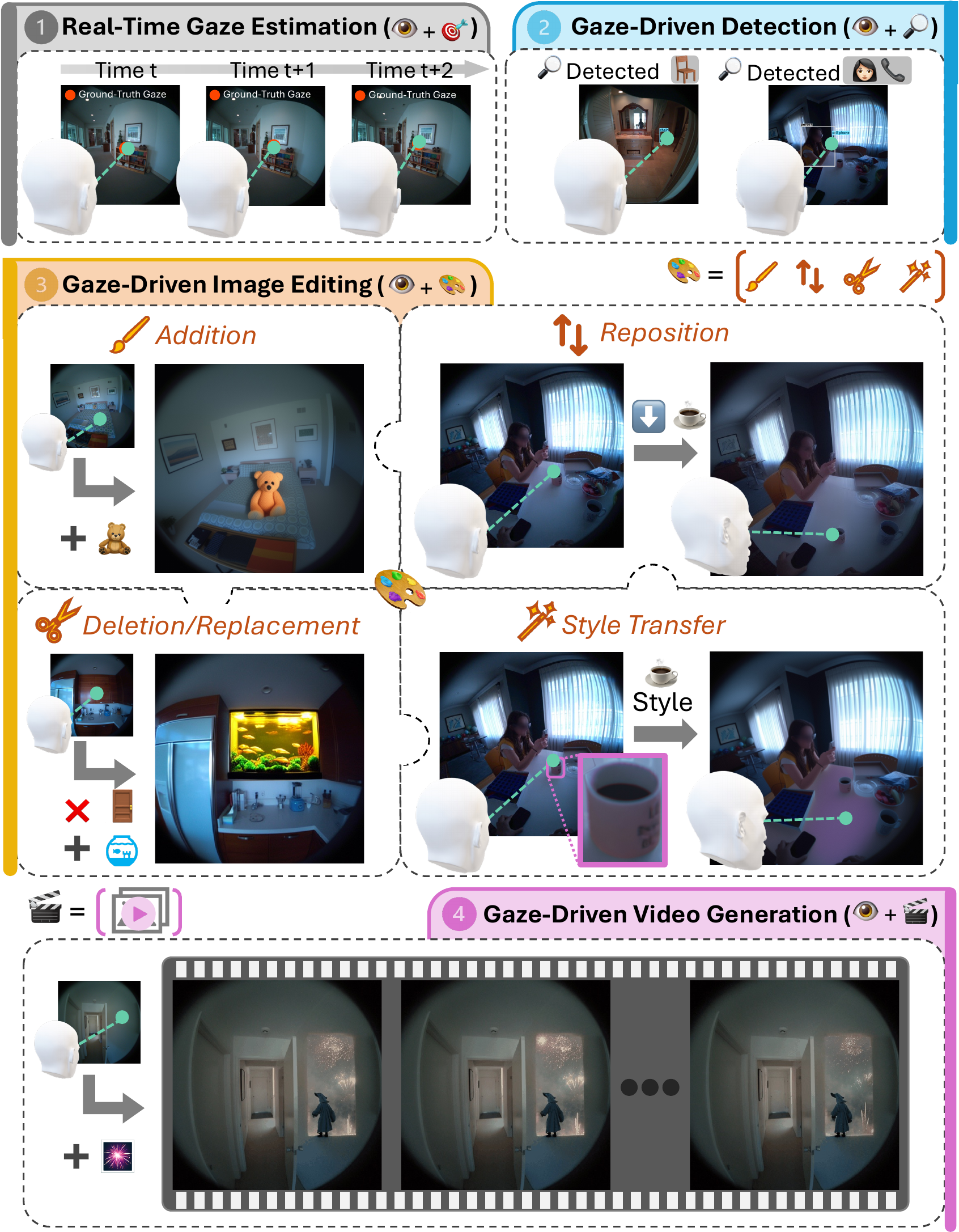}
    \captionof{figure}{
    \hy{
    Extended applications of gaze-driven interaction with GazeGen.
    (1) \textit{Real-Time Gaze Estimation}: Continuous tracking of eye movements for precise gaze estimation.
    (2) \textit{Gaze-Driven Detection}: Detecting and identifying objects based on where the user is looking.
    (3) \textit{Gaze-Driven Image Editing}: Dynamic editing tasks such as \textit{Addition} (adding objects based on the user's gaze), \textit{Deletion/Replacement} (removing or replacing objects based on the user's gaze), \textit{Reposition} (move objects by first gazing at the initial position, then the new position), and \textit{Style Transfer} (change an object's style or texture by first gazing at a reference object, then applying the style to the target object).
    (4) \textit{Gaze-Driven Video Generation}: Creating and manipulating video content driven by the user's gaze.
    }
    }
    \label{fig:gazeApps}
\end{figure*}

\section{Introduction}
\hy{
Recent advancements in visual content editing interfaces have highlighted the need for systems that are both intuitive and accessible. Traditional methods often rely on physical manipulation, which can be limiting, especially for individuals with physical disabilities. To address this, we introduce GazeGen, a system leveraging eye gaze for hands-free interaction, enhancing user engagement and accessibility beyond conventional augmented reality (AR) environments. By utilizing natural human behavior—where gaze directs attention and guides actions—GazeGen provides a straightforward interface for managing and interacting with digital content. This approach capitalizes on instinctual behaviors, such as looking and seeing, to simplify complex operations, making GazeGen more user-friendly and widely accessible.
\\
Consider a designer adjusting visual elements in a digital design platform. Traditionally, this task requires manual adjustments, which can be cumbersome and time-consuming. With GazeGen, the designer simply looks at the elements they want to adjust. The system interprets these gaze points as commands, enabling immediate and precise edits. Real-time eye interaction is crucial as it allows for seamless and intuitive control, and since everyone has different eye shapes and movements  \cite{HePVXRLN19, KrafkaKKKBMT16, ZhangHSB18, YuLO19, ParkMMIHK19, LindenSP19, LiuYMO18, LiuYMO21, ChenS20, LiuQLHWPY24}, personalization is essential for accuracy. This capability not only accelerates the creative process but also makes it more inclusive, allowing anyone to express their creativity regardless of physical capabilities.
\\
At the core of GazeGen is the DFT Gaze (Distilled and Fine-Tuned Gaze) agent, an ultra-lightweight gaze estimation model designed for real-time, accurate predictions tailored to individual users. DFT Gaze captures gaze points in real time for both object retrieval and visual content manipulation. Integrating gaze estimation technology into visual content generation applications presents unique challenges, which GazeGen addresses through effective personalization for accurate gaze prediction and a lightweight design.
The DFT Gaze agent is designed to be adaptable and efficient, requiring minimal computational resources for real-time interactions. It learns from general gaze patterns and supports easy personalization with just a few user-specific samples. With only 281K parameters, DFT Gaze is very compact, achieving performance comparable to larger models while operating 2x faster on edge devices (\eg, the Raspberry Pi 4).
The lightweight and real-time capabilities of DFT Gaze enable direct manipulation of objects through eye gaze. This allows users to interact with digital content naturally and intuitively, enabling hands-free interactions in AR environments. We demonstrate the broad applications of GazeGen in Fig. \ref{fig:gazeApps}.
\\
With advanced object detection and generative AI methods, GazeGen extends the functionality of eye gaze from simple tracking to dynamic visual content manipulation. Users can perform complex tasks such as adding, deleting, repositioning elements, and even transforming static images into videos, all through their gaze. This capability makes visual content creation accessible to everyone, regardless of physical limitations, and enhances the creative process with a seamless, unobtrusive interface.
\\
To support these advanced functionalities, we begin by developing a compact gaze estimation model through knowledge distillation. This process preserves the teacher model's knowledge while significantly reducing computational complexity by reconstructing the teacher's features using self-supervised learning. To achieve accurate gaze estimation, we integrate Adapters into this model, allowing it to learn diverse gaze patterns and personalize predictions for individual users.
\\
With this robust gaze estimation foundation, GazeGen extends its capabilities to real-time object detection by leveraging gaze points to focus on specific regions of the image, retrieving object categories and bounding boxes. For visual content generation, GazeGen uses gaze as a natural command for dynamic image editing and video creation, enabling intuitive operations such as addition, deletion, repositioning, and style transfer. This comprehensive approach allows users to seamlessly manipulate visual content through their gaze, setting a new standard for accessibility and efficiency in the field.
\\
GazeGen offers a new standard in gaze-driven visual content generation with the following key contributions:
\begin{enumerate}
    \item \textbf{Use of Eye Gaze for Visual Content Manipulation}: We are the first to propose using eye gaze for comprehensive visual content generation and editing, such as adding, deleting, repositioning elements, style transfer, and generating videos. Additionally, GazeGen can detect and interact with objects based on where the user is looking, offering a hands-free and intuitive interface for content manipulation.
    \item \textbf{Compact and Efficient Gaze Model}: We developed the DFT Gaze agent, a highly compact gaze estimation model with only 281K parameters, created through knowledge distillation coupled with a masked autoencoder. Our model leverages self-supervised learning techniques to reconstruct input images and teacher network features, effectively capturing the teacher's knowledge. Despite its compact size, the student model shows minimal performance drop compared to the teacher model and achieves 2x faster performance on the edge device.
    \item \textbf{Enhanced User Experience}: GazeGen leverages natural human behaviors, providing a seamless and intuitive interface for visual content manipulation. By personalizing gaze estimation with minimal samples, our system adapts to individual users, ensuring high accuracy and ease of use.
    \item \textbf{Broad Application Scope}: We demonstrate the wide applicability of GazeGen in various scenarios. Fig. \ref{fig:gazeApps} illustrates the diverse potential applications of our system \footnote{Text can be converted through voice.}, showcasing its versatility and effectiveness.
\end{enumerate}
}

\section{Preliminary}
\hy{
This section details the key components of GazeGen: Knowledge Distillation (KD), Adapters, and Stable Diffusion (SD). These components are foundational for advancing gaze-driven interaction. The DFT Gaze model, designed for precise gaze estimation, employs KD and Adapters to achieve high accuracy. Integrated with SD, the DFT Gaze model constitutes the core of GazeGen, facilitating sophisticated visual editing and interaction capabilities.
\\
\textbf{Knowledge Distillation (KD):} Knowledge Distillation transfers knowledge from a large, complex neural network (the teacher) to a smaller, more efficient one (the student). This process allows the student model to perform nearly as well as the teacher with significantly less computational power. In our system, feature-based knowledge distillation is employed to enhance the student model by aligning its visual processing abilities with those of the teacher model. This alignment involves minimizing the discrepancies in how both models process and interpret visual information, ensuring that the student model not only retains but effectively utilizes the high-level insights learned by the teacher.
\\
\textbf{Adapters:} Adapters are compact modules added to pre-trained neural networks to enable fine-tuning for specific tasks without the need to retrain the entire model. By applying a simple transformation:
\[
\text{feature}_{\text{new}} = \text{feature}_{\text{original}} + \text{Adapter}(\text{feature}_{\text{original}}),
\]
where \( \text{feature}_{\text{original}} \) represents the feature vector produced by the standard layers of the model, and \( \text{Adapter}(\cdot) \) is the function implemented by the adapter module. Adapters adjust the model's output, enhancing its task-specific performance while preserving the original network architecture. This method is efficient, leveraging pre-trained weights that already encode valuable general knowledge, thus avoiding the costly process of training from scratch.
In the DFT Gaze model, adapters are introduced post knowledge distillation to fine-tune generic and personalized gaze patterns. This adaptation significantly enhances gaze estimation accuracy by tailoring the model to individual user characteristics.
\\
\textbf{Stable Diffusion (SD):}
Stable Diffusion (SD) serves as a generative engine to transform textual descriptions into visual content, specifically Text-to-Image (T2I) and Text-to-Video (T2V), valued for its flexibility and strong community support. It begins by encoding an image into a latent representation \( z_0 = \mathcal{E}(x_0) \) within a pre-trained autoencoder's latent space. 
\\
The transformation process involves modifying \( z_0 \) through a series of diffusion steps:
\[
z_t = \sqrt{\alpha_t}z_0 + \sqrt{1-\alpha_t}\epsilon, \quad \epsilon \sim \mathcal{N}(0, I),
\]
for each step \( t \), where \( \alpha_t \) controls the noise level. The denoising model \( \theta(\cdot) \) works to reverse these additions and restore the image using the textual prompt \( y \) and the text encoder \( \tau(\cdot) \).
\\
The network architecture of \( \theta(\cdot) \) features a U-Net structure optimized for various resolution levels, integrating ResNet blocks, spatial self-attention, and cross-attention mechanisms to respond adaptively to the textual prompts in the image synthesis.
\\
Leveraging prior knowledge from generative models, GazeGen generates and edits high-quality visual content directed by user gaze, operating without the need for dataset fine-tuning. By interpreting gaze points as commands for precise edits, this method simplifies the intricate and labor-intensive nature of graphic design tasks. GazeGen accelerates the creative process and enhances inclusivity, allowing anyone to express their creativity regardless of physical capabilities.
}

\section{GazeGen}

\begin{figure}[!t]
    \centering
    \includegraphics[width=0.7\linewidth]{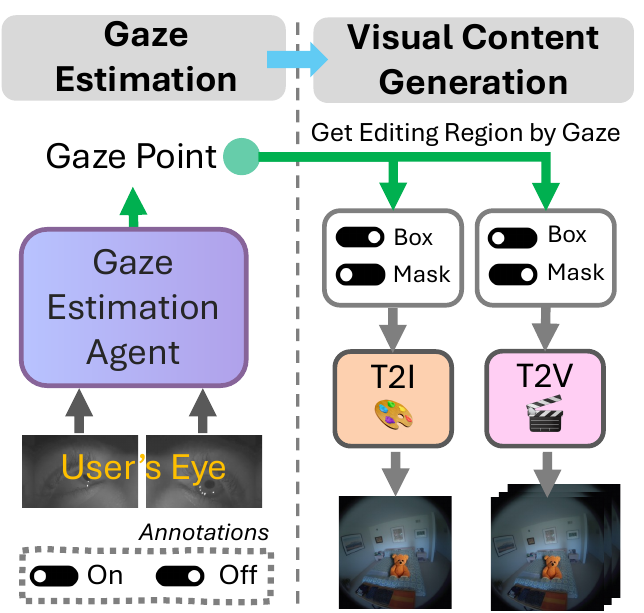}
    \caption{
    \hy{
    Gaze-driven visual content generation. This diagram shows the process starting from the user's eye, where the gaze estimation agent determines the gaze point. The gaze point is used to get the editing region, which can be toggled to use either a box or a mask. The T2I (Text-to-Image) and T2V (Text-to-Video) modules then generate visual content based on the selected editing region. The On/Off switches indicate whether the box or mask is used for gaze-driven editing.
    }
    }
    \label{fig:gaze2sd}
\end{figure}

\hy{
GazeGen enhances user interaction by leveraging eye-gaze to generate and edit visual content. As shown in Fig. \ref{fig:gaze2sd}, the system integrates a gaze estimation agent with visual content generation techniques, dynamically adapting to the user's gaze patterns. First, in Sec. \ref{sec:kd}, we reduce the larger, complex ConvNeXt V2 Atto (ConvNeXt V2-A) network into a more compact yet effective model capable of capturing essential visual details. Next, in Sec. \ref{sec:adapter}, we enhance this compact model, now referred to as DFT Gaze, with Adapters to better align with individual gaze patterns. Finally, in Sec. \ref{sec:detection} and \ref{sec:generation}, we utilize gaze predictions from the real-time gaze estimation model to dynamically detect objects and generate and modify visual content. Detailed explanations of the training and operational mechanisms of GazeGen are provided in Sec. \ref{sec:inpractice}.
}

\begin{figure}[!t]
    \centering
    \includegraphics[width=0.9\linewidth]{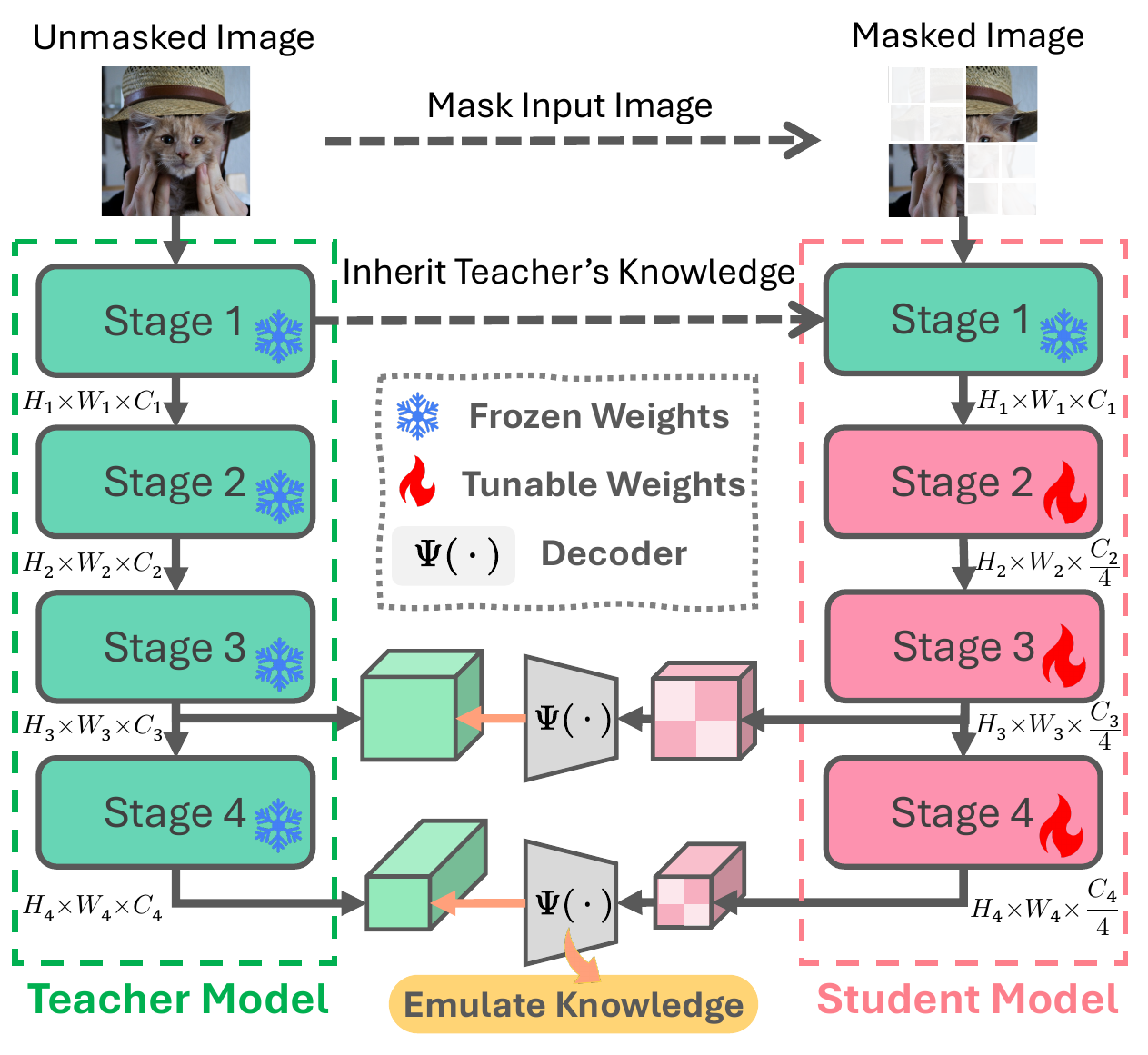}
    \caption{
    \hy{
    Self-supervised distillation for a compact model. Using ConvNeXt V2-A \cite{WooDHC0KX23} as the teacher network, we create a downsized student network. The first stage of the student model inherits weights from the teacher, while stages 2 to 4 reduce the channel dimensions to one-fourth. Distinct decoders are used to reconstruct both input images and the teacher’s intermediate features. The student processes masked inputs, allowing it to emulate the teacher's deep understanding of visual data and align with how the teacher perceives and interprets these images. For simplicity, the diagram only illustrates the reconstruction of the teacher’s features to emulate knowledge.
    }
    }
    \label{fig:kd}
\end{figure}

\hy{
\subsection{Self-Supervised Compact Model Distillation}
\label{sec:kd}
Efficient gaze estimation is fundamental for GazeGen, given the computationally intensive tasks of visual content generation and object retrieval. These tasks necessitate an exceptionally fast and precise gaze estimation model to minimize overall latency. To address this, we developed a compact model that effectively balances speed and precision, essential for facilitating smooth user interactions.
Using the ConvNeXt V2-A \cite{WooDHC0KX23} framework, known for its high performance in image classification and low overhead, we applied knowledge distillation to create a \textit{student model}. This streamlined version of the more complex \textit{teacher model} (ConvNeXt V2-A) maintains the ability to process complex visual information effectively. The student model adopts the architecture of the teacher but with reduced complexity by reducing the channel dimensions to one-fourth in each ConvNeXt V2 Block, as depicted in Fig. \ref{fig:kd}.
\\
In the knowledge distillation phase, the student model processes masked input images from ImageNet-1K \cite{DengDSLL009}, aiming to reconstruct both the original images $\mathcal{X}$ and the teacher’s intermediate features $\mathbf{f}^T$. This approach allows the student model to emulate the teacher's deep understanding of visual data, aligning with how the teacher perceives and interprets these images.
\\
We specifically focus on reconstructing high-level features in the last two stages ($l$-th stage, where $l \in \{3, 4\}$) of the ConvNeXt V2-A, while the first stage uses the same weights as the teacher. This setup ensures that the student model builds on the same fundamental knowledge, allowing it to develop and process abstract concepts similarly. The dual reconstruction tasks, aligning on how data is represented and perceived, help the student model closely match the teacher’s advanced capabilities, even with partial inputs.
\\
Each reconstruction task is handled by a distinct ConvNeXt V2 Block \cite{WooDHC0KX23} acting as a decoder, tailored to manage both image and feature reconstructions efficiently. To reconstruct the intermediate features from the teacher network, we express the decoder $\Psi(\mathbf{z})$ with an input $\mathbf{z}$ as:
\[
\Psi(\mathbf{z}) = \mathrm{FC}(\mathbf{z} + \mathrm{Conv}_{1\times1}(\mathrm{GRN}(\mathrm{GELU}(\mathbf{\hat{z}}))))
\]
where $\mathbf{\hat{z}} = \mathrm{Conv}{1\times1}(\mathrm{LN}(\mathrm{DConv}{7\times7}(\mathbf{z})))$. We align the student's features, $\mathbf{f}^S_l$, with those of the teacher, $\mathbf{f}^T_l$, at the same stage using this decoder.
The reconstruction loss, which considers both the input image and intermediate features, is defined as:
\begin{flalign}
\label{eq:recons}
\begin{aligned}
\mathcal{L}_{recon} &= \frac{1}{\phi(\mathcal{X}_K)} \sum_{k \in K} (\mathcal{X}_k - \hat{\mathcal{X}}_k)^2 + \\ &
\gamma \sum_{l \in \{3, 4\}} \frac{1}{\phi(\mathbf{f}^T_{l,K})} \sum_{k \in K} (\mathbf{f}^T_{l,k} - \Psi(\mathbf{f}^S_{l,k}))^2,
\end{aligned}
\end{flalign}
where $K$ represents the set of masked pixels in both the original images and the corresponding feature maps. $\phi(\cdot)$ denotes the count of these pixels in each context, and $\gamma = 0.5$ balances the loss components between image and feature reconstruction.
}

\hy{
\subsection{Gaze Estimation Interpreting with Adapters}
\label{sec:adapter}
To achieve accurate gaze estimation tailored to individual users, we enhance the streamlined model developed through knowledge distillation by integrating Adapters, transforming it into the DFT Gaze model. This adaptation serves two key purposes: 1) to learn from a comprehensive dataset that captures a wide range of gaze patterns from various participants, and 2) to tailor the model to the unique gaze dynamics of each user, which is critical due to individual variations in eye anatomy and gaze behavior.
\\
\textbf{Generalized Gaze Estimation.} 
In the generalized phase, the DFT Gaze model uses Adapters, each consisting of two fully-connected (FC) layers with BatchNorm (BN) and LeakyReLU (LReLU) activation, to learn gaze variations. These Adapters are specifically fine-tuned to improve responsiveness to varied gaze data, while the rest of the model remains unchanged to leverage existing visual knowledge. 
\\
The training involves a generalized dataset ($D_G$) containing gaze data from all participants, which is clustered into 15 groups using K-means to address imbalances in gaze direction distributions. This clustered generalized dataset ($\mathcal{G}$) ensures that the model learns from a balanced and comprehensive representation of diverse gaze behaviors, facilitating a more uniform adaptation to different gaze patterns.
\\
To adaptively adjust gaze features within the DFT Gaze model, Adapters are applied to modify internal features in each ConvNeXt V2 Block. The transformation is defined as:
\[
\mathrm{Adapter}(\mathbf{f}^{V}) = \mathrm{FC_{up}}(
        \mathrm{LReLU}(
            \mathrm{BN}(\mathrm{FC}_{down}(\mathbf{f}^{V})
        )))
\]
Here, $\mathbf{f}^{V}$ denotes the features from the final Convolutional layer of each block. The $\mathrm{FC}_{down}$ layer initially compresses these features to a quarter of their original dimension, isolating the most crucial attributes. This compression simplifies the feature space, enhancing focus during learning. Subsequently, the $\mathrm{FC}_{up}$ layer restores the features to their original dimensions, allowing the model to integrate these refined features while maintaining the overall structure of the feature space.
\\
\textbf{Personalized Gaze Estimation.} 
Following the generalized phase, personalization is essential to adapt the model to each user's unique gaze dynamics, considering individual differences in eye anatomy and behavior. The personalization phase focuses on fine-tuning the Adapters in the final stage of the DFT Gaze model. This fine-tuning uses a personalized dataset ($D_P$) comprising only five personal eye gaze images per participant. To prevent overfitting and maintain the model’s generalization capabilities, known as avoiding \textit{model forgetting} \cite{LangeAMPJLST19, RuizLJPRA23, ParkMMIHK19, SchneiderV21, LiuQLHWPY24}, we reintroduce a subset of the clustered generalized dataset ($\mathcal{G}$) during this phase. This approach preserves the model’s robustness across diverse gaze patterns and enhances its precision for personalized gaze estimation, resulting in high accuracy for individual-specific scenarios. Table \ref{tab:std_tchr_comp} presents the low angular gaze error achieved with DFT Gaze on both the AEA and OpenEDS2020 datasets.
}

\hy{
\subsection{Gaze-Driven Object Detection}
\label{sec:detection}
Having established a fast and accurate gaze estimation model, we extended its capabilities to recognize objects users are looking at. Our approach to object detection is training-free and leverages gaze points to streamline the process. While existing object detectors \cite{JocherCQ23, WangYL24} analyze the entire feature map by considering all grid cells to predict objects' coordinates and classes, our method specifically focuses on the area around the gaze point. The gaze point, represented as a 2-dimensional coordinate \((x, y)\), is used to retrieve the relevant feature grid cells and their neighboring cells, each corresponding to a specific region of the original image. This method reduces computational overhead and accelerates detection by concentrating only on these gaze-directed grid cells.
\\
Specifically, let \( G \) represent the feature grid, and \( g_{i,j} \) the grid cell at position \((i, j)\). Given the gaze point \((x, y)\) in the image space, we identify the corresponding grid cell \( g_{x,y} \) and its neighboring cells within a certain range \( k \). This range includes cells \( g_{x \pm m, y \pm n} \) for \( m, n \in \{0, 1, 2, \ldots, k\} \). The object detection is then focused on these cells $\{g_{x+m, y+n} \mid m, n \in \{-k, \ldots, -1, 0, 1, \ldots, k\}\}$. By targeting this set of specific cells, we efficiently predict the bounding boxes and classes relevant to the user's focus, optimizing detection based on real-time gaze input. This approach can further reduce the processing time of non-maximum suppression.
}

\hy{
\subsection{Gaze-Driven Visual Content Generation}
\label{sec:generation}
Beyond simply recognizing objects users are looking at, we ask: \textit{Can we create and edit visual content using just our eyes}? GazeGen enables dynamic visual content generation and editing, leveraging gaze as a natural command. This makes the process more efficient and closely aligned with user intentions. GazeGen incorporates both gaze-driven image editing and video generation, utilizing forward diffusion and reverse diffusion.
%
\paragraph{Gaze-Driven Image Editing}
We introduce gaze-driven operations such as Addition, Deletion/Replacement, Repositioning, and Style Transfer, facilitating intuitive editing of visual content, gaining insights from recent advancements in image editing.
\\
%
\textbf{Addition.} To incorporate new objects based on the user's gaze, we use MLLM (\eg, LLaVA \cite{LiuLWL23a}) suggests the bounding box from the user's gaze point, and generative AI synthesizes the object within the specified area.
\\
\textbf{Deletion/Replacement.} For deletion, the object area is removed, and generative AI regenerates the region to ensure a coherent image. For replacement, generative AI synthesizes a new object within the same area.
\\
\textbf{Repositioning.} Repositioning is achieved by tracking multiple gaze points to determine the new position for an object. The object is moved to its new location, and the original area is filled to ensure a consistent background. Generative AI refines the object and blends it into its new surroundings.
\\
\textbf{Style Transfer.} The process uses eye gaze to guide the extraction and transfer of style onto a target object using generative AI.
\paragraph{Gaze-Driven Video Generation}
We extend Text-to-Video (T2V) models, by transforming a user's viewed image into animation. Using gaze coupled with LLaVA to suggest bounding boxes and add animated objects, we edit and animate visual content based on user gaze. This integration enables intuitive and dynamic video creation, where the user's gaze directs the animation process, allowing for interactive video generation.
\\
\textbf{Addition.} 
To incorporate animated objects into a T2V model using gaze, reverse diffusion is employed to generate cohesive and dynamic animations.
\\
\textbf{Replacement.}
For replacement, the object's area is removed, and generative AI synthesizes an animated object.
}

\hy{
\subsection{GazeGen in Practice}
\label{sec:inpractice}
All experiments were conducted on a desktop with an Intel Core i9-13900K CPU and an Nvidia GeForce RTX 4090 GPU.
\\
\textbf{Self-Supervised Compact Model Distillation.}
We perform knowledge distillation through self-supervised learning on the ImageNet-1K dataset \cite{DengDSLL009}. ConvNeXt V2 Atto \cite{WooDHC0KX23} serves as the teacher network, utilizing the officially released checkpoint\footnote{\scriptsize{\url{https://github.com/facebookresearch/ConvNeXt-V2}}}. The reconstruction loss is calculated using Eq. (\ref{eq:recons}).
\\
\textbf{Gaze Estimation Interpreting with Adapters.}
We use L1 loss to minimize gaze prediction errors and report mean angular gaze error following prior studies \cite{PalmeroSBKKT20, ZhangPBBTH20, CaiZS023}. In the generalized phase, a generalized dataset ($D_G$) is clustered into 15 groups using K-means to balance gaze direction distributions, forming the clustered dataset ($\mathcal{G}$). In the personalized phase, a personalized dataset ($D_P$) with 5 personal eye gaze images per participant is supplemented by a subset of $\mathcal{G}$ to avoid model forgetting.
\\
\textbf{Gaze-Driven Visual Content Generation/Detection.}
We leverage advanced models to achieve training-free gaze-driven visual content generation and detection, enabling intuitive user interactions. For image editing, objects are added based on bounding boxes suggested by LLaVA \cite{LiuLWL23a}. Additionally, YOLOv9 \cite{WangYL24} identifies and classifies objects within the scene, facilitating gaze-driven object detection.
}

\section{Experiments}

\hy{
\subsection{Dataset Details}
\textbf{OpenEDS2020.} The OpenEDS2020 dataset \cite{PalmeroSBKKT20} is a 3D gaze estimation dataset of eye images collected using a VR head-mounted device. For generalized gaze estimation, we used the training set as the generalized set ($D_G$) and evaluated the model on the validation set. For personalized gaze estimation, the testing set was used, with each participant providing only 5 images for fine-tuning and the remaining images for evaluation. We reported the average angular gaze error over all participants.
\\
\textbf{AEA (Aria Everyday Activities) Dataset.} The AEA dataset \cite{LvC24} consists of eye images captured during various daily activities, providing a diverse range of gaze scenarios. This dataset includes images collected in natural settings, offering a realistic environment for gaze estimation. We partitioned the data with a 8:1:1 ratio to create the generalized training set ($D_G$), generalized test set, and personalized set ($D_P$). Five images per participant were selected for personal fine-tuning. The generalized model was trained on $D_G$ and evaluated on the generalized test set. The personalized model was fine-tuned on $D_P$ and then evaluated on the remaining images.
\\
\textbf{Clustering and Fine-Tuning.} For both datasets, K-means clustering with $K=15$ was applied to $D_G$ to build a clustered generalized set ($\mathcal{G}$). During the fine-tuning of the personalized model, a small subset of $\mathcal{G}$ was used to avoid model forgetting.
\\
\textbf{Evaluation Metrics.} Following prior studies \cite{ParkSH18, ParkMMIHK19, PalmeroSBKKT20, ZhangPBBTH20, CaiZS023}, we report the mean angular gaze error (in $^\circ$) for the gaze estimation task.
}

\hy{
\subsection{Teacher-Student Model Comparison}
We evaluated the performance of our gaze estimation models using both generalized and personalized datasets to compare the teacher model, ConvNeXt V2-A, with the student model, DFT Gaze, as shown in Tab. \ref{tab:std_tchr_comp}.
\\
\textbf{Generalized Gaze Estimation.} The ConvNeXt V2-A model, with 3.6M parameters, achieved a mean angular error of 1.94$^\circ$ on the AEA dataset and 6.90$^\circ$ on the OpenEDS2020 dataset. The DFT Gaze model, significantly smaller with 281K parameters, demonstrated a slightly higher mean angular error of 2.14$^\circ$ on the AEA dataset and 7.82$^\circ$ on the OpenEDS2020 dataset. Despite the reduced number of parameters, the student model maintained competitive performance, highlighting its efficiency.
\\
\textbf{Personalized Gaze Estimation.} For personalized gaze estimation, the ConvNeXt V2-A model achieved a mean angular error of 2.32$^\circ$ on the AEA dataset and 5.36$^\circ$ on the OpenEDS2020 dataset. The DFT Gaze model, with its compact size, achieved a mean angular error of 2.60$^\circ$ on the AEA dataset and 5.80$^\circ$ on the OpenEDS2020 dataset. The minimal performance drop demonstrates the robustness of the student model in personalized settings.
}

\begin{figure}[h]
    \renewcommand{\numColumns}{2}
    \begin{tabular}{
        @{}
        p{\dimexpr(\columnwidth-\columnSpacing*(\numColumns-1))/\numColumns} @{\hspace{\columnSpacing}}
        p{\dimexpr(\columnwidth-\columnSpacing*(\numColumns-1))/\numColumns} 
        @{}
    }
        \animategraphics[width=0.45\columnwidth]{8}{videos/track/l2s4s7r1/}{01}{16} & 
        \animategraphics[width=0.45\columnwidth]{8}{videos/dets/l1s2s6r1/}{01}{16}
    \end{tabular}
    \begin{tabularx}{\columnwidth}{XXX}
        \emph{\small \textbf{(Eye Tracking)}} & 
        \emph{\small \textbf{(Object Detection)}}
    \end{tabularx}
        \caption{
        \hy{
        Qualitative results on \aea~dataset. First row: user’s eye. Second row: eye tracking (left) and gaze-driven object detection (right). Predicted gaze (\textcolor{ForestGreen}{green}), ground-truth gaze (\textcolor{Red}{red}). \animationNotes
        }
    }
    \vspace{1em}
    \label{fig:eye_track_det}
\end{figure}

\vspace{-5mm}
\begin{figure*}
    \centering
    \includegraphics[width=0.99\linewidth]{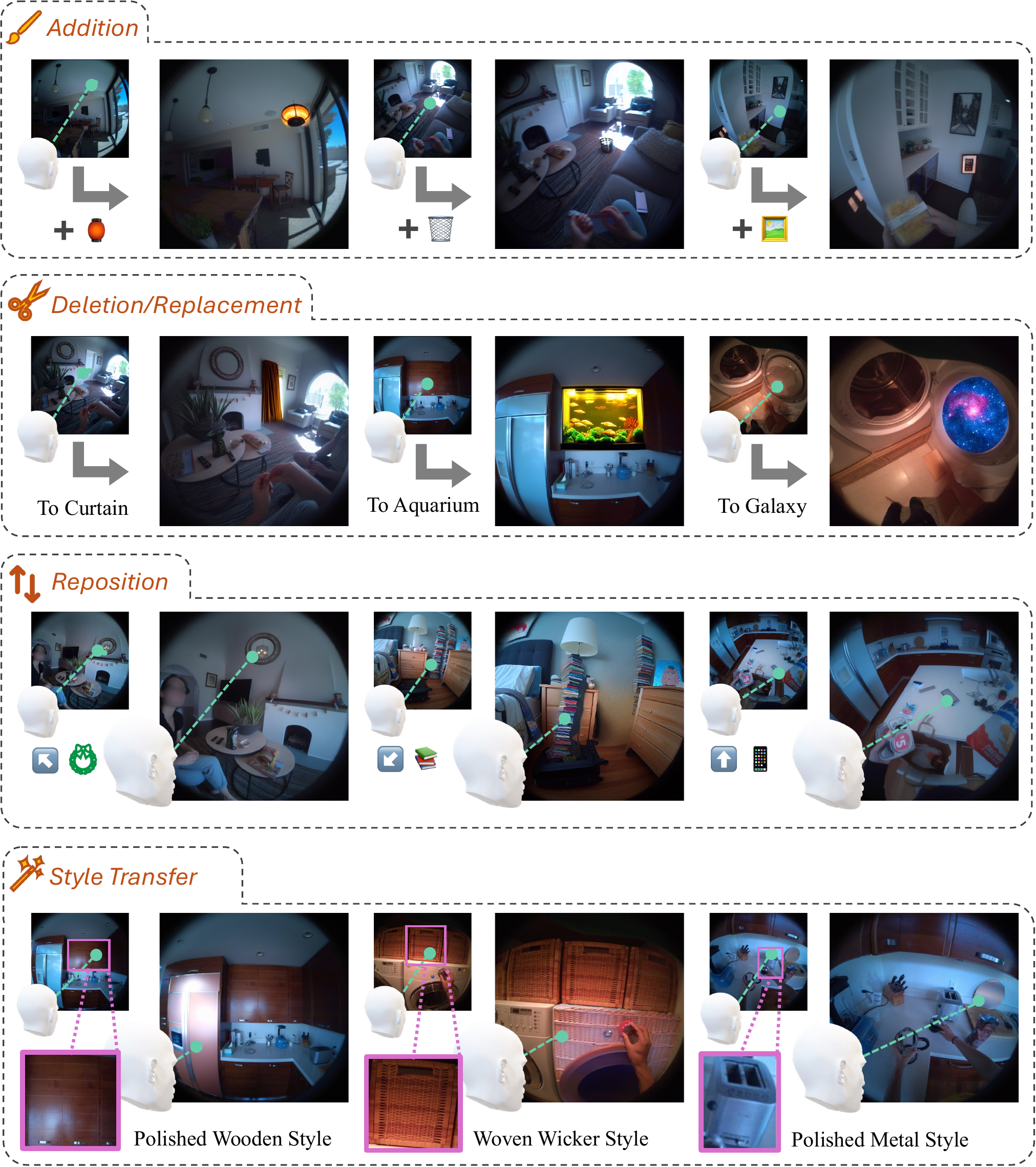}
    \captionof{figure}{
    \hy{
    Qualitative results for gaze-driven image editing. The tasks include: \textit{Addition} (first row): Adding objects like a lantern, basket, or photo. \textit{Deletion/Replacement} (second row): Replacing objects with items like a curtain, aquarium, or galaxy. \textit{Reposition} (third row): Moving objects such as a wall decoration to the upper left corner, books to the lower left corner, or a phone upward. \textit{Style Transfer} (last row): Changing an object's style, such as polished wood to the fridge, woven wicker to the washing machine, or polished metal to the chopping board. All edits are based on the user's gaze.
    }
    }
    \label{fig:imgApps}
\end{figure*}

\begin{figure*}[h]
    \renewcommand{\numColumns}{3}
    \begin{tabular}{
        @{}
        p{\dimexpr(\textwidth-\columnSpacing*(\numColumns-1))/\numColumns} @{\hspace{\columnSpacing}}
        p{\dimexpr(\textwidth-\columnSpacing*(\numColumns-1))/\numColumns} @{\hspace{\columnSpacing}}
        p{\dimexpr(\textwidth-\columnSpacing*(\numColumns-1))/\numColumns} @{}
    }
        \animategraphics[width=\linewidth]{8}{videos/animation/l5s4s2r1_00585/}{01}{16} & 
        \animategraphics[width=\linewidth]{8}{videos/animation/l5s5s1r1_00283/}{01}{16} &
        \animategraphics[width=\linewidth]{8}{videos/animation/l5s4s2r1_00111/}{01}{16}
    \end{tabular}
    \begin{tabularx}{\textwidth}{XXX}
        \emph{\small \textbf{A serene river flows gently with sparkling waves, with stones visible under the water}} & 
        \emph{\small \textbf{A night sky filled with twinkling stars}} &
        \emph{\small \textbf{A vibrant aquarium with fish swimming gracefully}}
    \end{tabularx}
        \caption{
        \hy{
        Qualitative results for gaze-driven video generation. Objects are replaced based on users' gaze with animated objects. \animationNotes \textit{Zoom in for a better view.}
        }
    }
    \vspace{1em}
    \label{fig:eye_video}
\end{figure*}

\begin{table*}[t]
\centering
\resizebox{0.99\textwidth}{!}{
\begin{tabular}{l c c | c c c c}
\multicolumn{1}{l}{Model} &
\multicolumn{1}{c}{\#param} &
\multicolumn{1}{c |}{tunable \#param} &
\multicolumn{1}{c}{\mpiigaze} &
\multicolumn{1}{c}{\mpiifacegaze} &
\multicolumn{1}{c}{\aea} & 
\multicolumn{1}{c}{\openeds}  
\\ 
\toprule
GazeNet \cite{ZhangSFB19} & 90.24M & 90.24M & 5.70 & 5.76 & 3.01 & 7.51\\
RT-Gene \cite{FischerCD18} & 31.67M & 31.67M & \textbf{4.61} & 4.66 & 2.03 & 6.02\\
GazeTR-Hybrid \cite{Cheng022} & 11.42M & 11.42M & - & \textbf{4.00} & \textbf{1.71} & \textbf{5.44}\\
\hline
ConvNeXt V2-A & 3.6M & 191.7K & 5.30 & 4.29 & 1.94 & 6.90\\
DFT Gaze & \textbf{281K} & \textbf{14.43K} & 6.13 & 5.17 & 2.14 & 7.82\\
\end{tabular}}
\caption{
\hy{
Comparison of state-of-the-art methods for generalized gaze estimation using within-dataset evaluation. To ensure a fair comparison, we reimplement these methods and apply the same K-means clustering with 15 groups as DFT Gaze during training. We follow the original hyperparameter settings specified in these methods.
}
}
\label{tab:sota_comp_general}
\end{table*}

\begin{table*}[t]
\centering
\resizebox{0.99\textwidth}{!}{
\begin{tabular}{l c c | c c c c}
\multicolumn{1}{l}{Model} &
\multicolumn{1}{c}{\#param} &
\multicolumn{1}{c |}{tunable \#param} &
\multicolumn{1}{c}{\mpiigaze} &
\multicolumn{1}{c}{\mpiifacegaze} &
\multicolumn{1}{c}{\aea} & 
\multicolumn{1}{c}{\openeds}  
\\ 
\toprule
GazeNet \cite{ZhangSFB19} & 90.24M & 90.24M & 5.39 & - & 4.16 & 6.57\\
RT-Gene \cite{FischerCD18} & 31.67M & 31.67M & - & 3.27 & 2.38 & 4.80\\
GazeTR-Hybrid \cite{Cheng022} & 11.42M & 11.42M & - & 3.04 & 2.05 & 3.43\\
$^\dagger$PNP-GA \cite{LiuLWL21} & 119.5M & 116.9M & - & 6.91 & - & -\\
$^\dagger$RUDA \cite{BaoLWL22} & 12.20M & 12.20M & - & 6.86 & - & -\\
$^\dagger$TPGaze \cite{LiuQLHWPY24} & 11.82M & 125K & - & 6.30 & - & -\\
\hline
ConvNeXt V2-A & 3.6M & 191.7K & 5.49 & 4.60 & 2.32 & 5.36\\
DFT Gaze & \textbf{281K} & \textbf{14.43K} & 6.61 & 5.35 & 2.60 & 5.80\\
\end{tabular}}
\caption{
\hy{
Comparison of state-of-the-art methods for personalized gaze estimation using within-dataset evaluation. To ensure a fair comparison, we reimplement these methods and apply the same K-means clustering with 15 groups as DFT Gaze during training. We follow the original hyperparameter settings specified in these methods. The symbol $\dagger$ represents source-free unsupervised domain adaptation (UDA) methods.
}
}
\label{tab:sota_comp_persl}
\end{table*}

\begin{table}[t]
\centering
\resizebox{0.95\columnwidth}{!}{
\begin{tabular}{l c c c c }
\multicolumn{1}{l}{Model} &
\multicolumn{1}{c}{\#param} &
\multicolumn{1}{c}{\aea} & 
\multicolumn{1}{c}{\openeds} \\ 
\toprule
\textbf{Generalized Gaze Estimation} \\
ConvNeXt V2-A (Teacher) & 3.6M & \textbf{1.94}
              & \textbf{6.90} \\
DFT Gaze (Student) & \textbf{281K} & 2.14
              & 7.82 \\
\hline
\textbf{Personalized Gaze Estimation} \\
ConvNeXt V2-A (Teacher) & 3.6M & \textbf{2.32}
              & \textbf{5.36} \\
DFT Gaze (Student) & \textbf{281K} & 2.60
              & 5.80 \\
\end{tabular}}
\caption{
\hy{
Generalized and personalized gaze Estimation results. The teacher model, ConvNeXt V2-A, with 3.6M parameters, excels in both generalization and personalization, achieving superior performance across all datasets. The student model, DFT Gaze, with only 281K parameters, shows minimal performance drop, maintaining competitive levels in both settings. Despite its compact size, the student model provides robust gaze estimation within a streamlined framework, demonstrating its efficiency and effectiveness.
}
}
\label{tab:std_tchr_comp}
\end{table}

\begin{figure}[!t]
    \centering
    \includegraphics[width=0.99\linewidth]{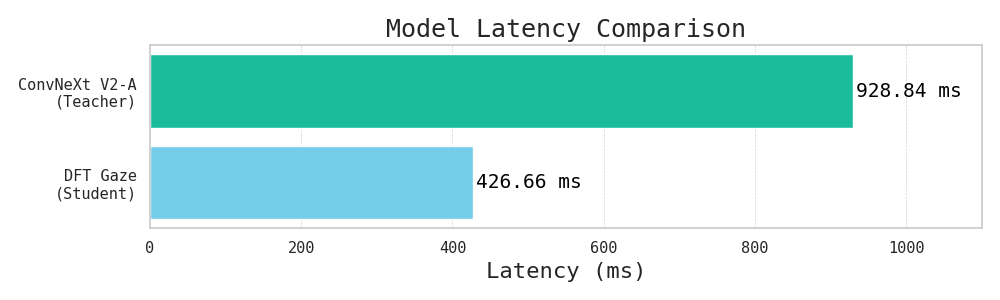}
    \caption{
    \hy{
    Model latency comparison on Raspberry Pi 4. The figure compares the latency of two gaze estimation models: ConvNeXt V2-A (Teacher) and DFT Gaze (Student). ConvNeXt V2-A shows a latency of 928.84 ms, while DFT Gaze reduces latency to 426.66 ms, demonstrating its efficiency for real-time applications on edge devices.
    }
    }
    \label{fig:latency}
\end{figure}

\hy{
\subsection{Gaze Estimation Latency on Edge Device}
To enable real-time gaze estimation for quick eye interaction and enhance user experience, which is crucial for subsequent visual content generation, we tested the latency of two models, ConvNeXt V2-A (teacher) and DFT Gaze (student), on a Raspberry Pi 4 with 8GB RAM. This widely-used edge device demonstrates the feasibility of deploying our model in real-world scenarios with limited computational resources. Using input eye images from the AEA dataset, we evaluated each model on 1,000 images. As shown in Fig. \ref{fig:latency}, ConvNeXt V2-A exhibits an average latency of 928.84 milliseconds (ms), while DFT Gaze reduces this to an average latency of 426.66 ms, making it more suitable for real-time applications on edge devices. Despite this latency reduction, DFT Gaze only shows a minor performance drop, as indicated in Table \ref{tab:std_tchr_comp}. In knowledge distillation (KD), we streamline the student model design while retaining rich visual knowledge from the teacher model. This process allows DFT Gaze to achieve significant latency reduction without substantial loss in accuracy, making it a practical solution for real-time gaze estimation on edge devices.
}

\subsection{Qualitative Results}
\hy{
In this section, we demonstrate the diverse applications of GazeGen, including real-time gaze estimation, gaze-driven detection, gaze-driven image editing, and gaze-driven video generation.
\\
\textbf{Real-Time Gaze Estimation and Detection.} We begin with real-time gaze estimation and gaze-driven object detection as shown in Fig.~\ref{fig:eye_track_det}. The first row displays the captured user’s eye movements. The second row presents eye tracking in real time on the left, while the right side illustrates how the system performs gaze-driven object detection, identifying one or multiple items based on the user’s gaze.
\\
\textbf{Gaze-Driven Image Editing.} Next, we present results from various gaze-driven image editing tasks as shown in Fig.~\ref{fig:imgApps}. \textit{Addition}: The first row shows how objects like a lantern, basket, or photo are added to the scene based on where the user looks, enhancing the environment interactively. \textit{Deletion/Replacement}: In the second row, objects are replaced or removed, such as swapping out items for a curtain, aquarium, or galaxy. This demonstrates the system's ability to dynamically transform the visual context. \textit{Reposition}: The third row illustrates repositioning, where objects like a wall decoration are moved to new locations, such as the upper left corner, books to the lower left corner, or a phone moved upward, all guided by the user's gaze. \textit{Style Transfer}: The final row demonstrates changing the style of objects based on the user's gaze. For instance, the style of the first object seen by the user is applied to the next object they look at. Examples include applying a polished wood texture to a fridge, woven wicker to a washing machine, or polished metal to a chopping board. These changes reflect how gaze can influence the aesthetic and functional attributes of objects.
\\
\textbf{Gaze-Driven Video Generation.} Lastly, we demonstrate gaze-driven video generation in Fig~\ref{fig:eye_video}, where static objects are replaced with other animated objects based on the user's gaze. This application highlights the dynamic and interactive nature of the system, making scenes more engaging as the user's focus changes.
}

\begin{figure}[ht]
    \centering
    \begin{minipage}{0.48\columnwidth}
        \centering
        \includegraphics[width=\textwidth]{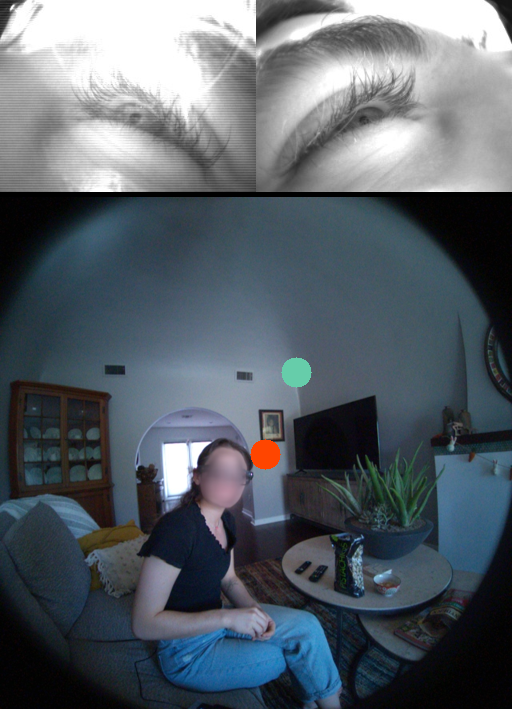} 
        \caption*{\small (a) Lighting conditions}
    \end{minipage}
    \hfill
    \begin{minipage}{0.48\columnwidth}
        \centering
        \includegraphics[width=\textwidth]{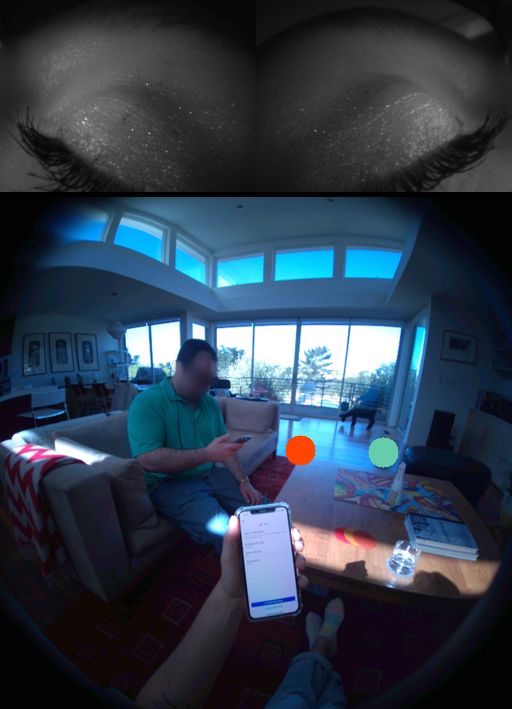} 
        \caption*{\small (b) Closed eyes}
    \end{minipage}
    \caption{
    Real-time gaze estimation limitations. The figure illustrates the DFT Gaze's limitations, showing deviations between predicted gaze (\textcolor{ForestGreen}{green}) and ground-truth gaze (\textcolor{Red}{red}) due to lighting conditions (left) and closed eyes (right). The top row shows users' eye images, while the bottom row visualizes the resultant gaze discrepancies.
    }
    \label{fig:tracking_limitations}
\end{figure}

\begin{figure}[ht]
    \centering
    \includegraphics[width=0.99\linewidth]{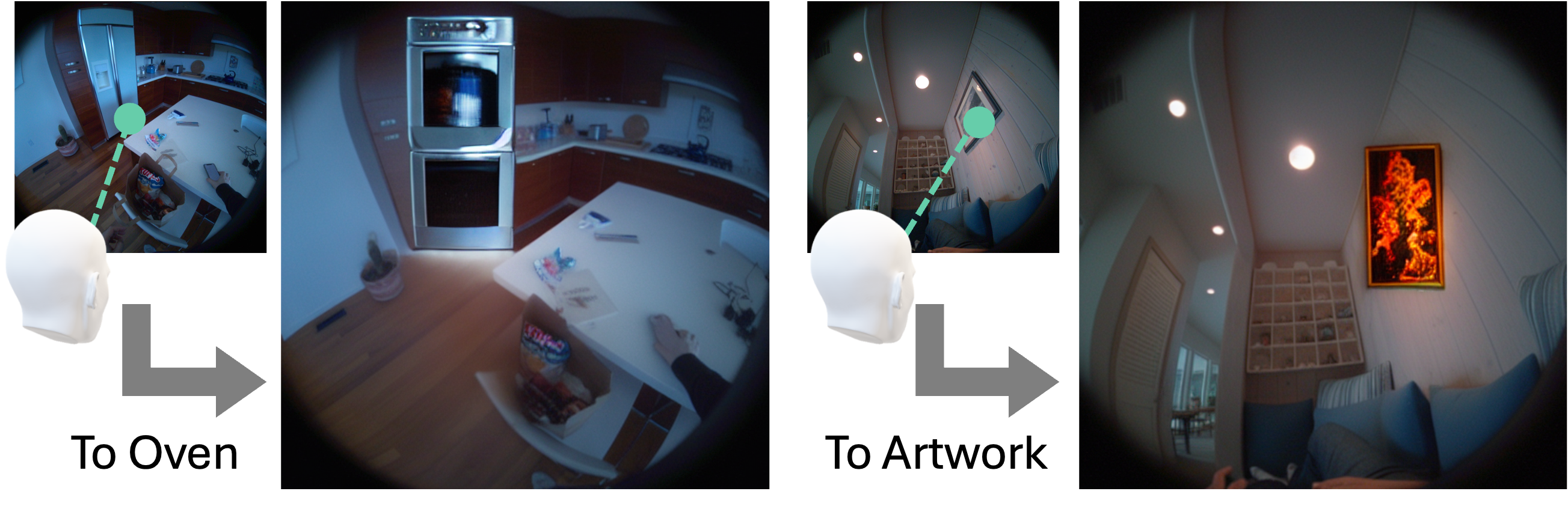}
    \caption{
    Visual content generation limitation. This figure illustrates the limitations of gaze-driven visual content generation. The replaced objects do not accurately reflect the original object's 3D angle or orientation, causing visual inconsistencies.
    }
    \label{fig:visual_limitations}
\end{figure}

\hy{
\section{Limitations }
\textbf{Real-Time Gaze Estimation Limitation.}
Despite DFT Gaze achieving accurate gaze predictions, it faces challenges under certain scenarios. These challenges primarily arise from:
(1) Lighting Conditions: Eye images often exhibit bright spots or glare due to reflective surfaces caused by lighting (see Fig. \ref{fig:tracking_limitations},  (a)). This can confuse the gaze estimation model, leading to errors in the predicted gaze. Implementing image preprocessing methods to remove glare and reflections could help mitigate this issue.
(2) Closed Eyes: When the user's eyes are closed, the gaze estimation model cannot provide accurate predictions (see Fig. \ref{fig:tracking_limitations},  (b)). The model relies on visible features such as the iris and pupil, which are not available when the eyes are closed. Considering previous eye images as hints could help avoid this limitation.
\\
\textbf{Visual Content Generation Limitation.}
Despite the effectiveness of gaze-driven visual content generation, the system still faces limitations. This figure, Fig.~\ref{fig:visual_limitations}, illustrates that the replaced objects do not accurately reflect the original object's 3D angle or orientation, causing visual inconsistencies. Enhancing the system to incorporate 3D modeling and perspective correction techniques could improve the accuracy of object replacements, potentially aligning them more closely with the original 3D angles and orientations. Additionally, implementing algorithms that address depth and spatial relationships could further refine the visual coherence of the generated content.
}

\hy{
\section{Related Work}
\textbf{Knowledge Distillation} is an effective compression technique that reduces model size by transferring knowledge from a deep network (teacher) to a lightweight network (student), enhancing inference speed while maintaining robust performance. Knowledge distillation can be categorized into logit distillation \cite{ZhangXHL18, FurlanelloLTIA18, ChoH19, MirzadehFLLMG20, ZhaoCSQL22} and intermediate representation distillation \cite{RomeroBKCGB14, KimPK18, HeoKYPK019, HeoLY019a, TianKI20, Bai0X0WYZX23}. Our method focuses on the latter, minimizing the difference between features from the teacher and student networks. FitNets \cite{RomeroBKCGB14} pioneered this approach by distilling intermediate representations. CRD \cite{TianKI20} uses contrastive learning to transfer structural data representations, while DMAE \cite{Bai0X0WYZX23} minimizes the distance between intermediate features using distinct architectures for teacher and student networks. Unlike DMAE, our method downsizes the teacher network's architecture to create the student network and transfers weights to its early stages, preserving detailed information. We then reconstruct the teacher network's features through decoders, ensuring the student model retains high-level insights learned by the teacher.
\\
\textbf{Personalized Gaze Estimation} \cite{HePVXRLN19, KrafkaKKKBMT16, ZhangHSB18, YuLO19, ParkMMIHK19, LindenSP19, LiuYMO18, LiuYMO21, ChenS20, LiuQLHWPY24} tailors gaze predictions to individual variations using a minimal set of personal gaze images, typically referred to as calibration points. This personalization enables precise mapping of gaze predictions to an individual's unique gaze patterns. In contrast, person-independent gaze models (referred to as generalized models in this paper) often yield low accuracies, exhibiting significant variability and person-dependent biases. For instance, SAGE \cite{HePVXRLN19} employs an unsupervised personalization approach for 2D gaze estimation, using unlabeled facial images and requiring at most five calibration points. Liu \etal~\cite{LiuYMO18, LiuYMO21} train a convolutional neural network to capture gaze differences between pairs of eye images, which is then used to predict the gaze direction for a new eye sample based on inferred differences. TPGaze \cite{LiuQLHWPY24} enhances personalization efficiency by updating a small set of parameters, termed "prompts," while keeping the network backbone fixed and employing meta-learning to optimize these prompts for adaptation.
}

\section{Conclusion}
This paper introduces GazeGen, a hands-free system for visual content generation using eye gaze, enhancing user engagement and accessibility in AR environments. At its core is the DFT Gaze agent, an ultra-lightweight model with 281K parameters, delivering real-time, accurate gaze predictions. It elevates eye gaze from basic tracking to dynamic visual manipulation, enabling tasks like adding, deleting, repositioning elements, style transfer, and converting static images into videos.
We developed a compact gaze estimation model using knowledge distillation and a masked autoencoder, refined with Adapters for precise, personalized gaze predictions. These predictions allow GazeGen to facilitate intuitive visual content manipulation and real-time object detection by targeting regions of interest indicated by the user's gaze, thus enhancing responsiveness and the creative process.
Overall, GazeGen sets a new standard for gaze-driven visual content generation, positioning users as active creators and broadening the scope of gaze-driven interfaces.

\bibliography{gazegen}

\end{document}